\documentclass{article}

\usepackage{times}
\usepackage{graphicx} 
\usepackage{subfigure}
\usepackage{mathtools}
\usepackage{amsfonts}

\usepackage{natbib}

\usepackage{algorithm}
\usepackage{algorithmic}

\usepackage{hyperref}

\usepackage[accepted]{icml2015}

\icmltitlerunning{T-SKIRT: Online Estimation of Student Proficiency in an Adaptive Learning System}

\begin{document}

\twocolumn[
\icmltitle{T-SKIRT: Online Estimation of Student Proficiency in an Adaptive Learning System}



\icmlauthor{Chaitanya Ekanadham}{chaitu@knewton.com}
\icmladdress{Knewton, Inc.,
           100 5th Avenue., New York, NY 10011 USA}
\icmlauthor{Yan Karklin}{yan@knewton.com}
\icmladdress{Knewton, Inc.,
           100 5th Avenue., New York, NY 10011 USA}

\icmlkeywords{item response theory, knowledge tracing, online learning, adaptive learning, student response prediction, proficiency estimation}

\vskip 0.3in
]

\begin{abstract}
We develop T-SKIRT: a temporal, structured-knowledge, IRT-based method for predicting student responses online. By explicitly accounting for student learning and employing a structured, multidimensional representation of student proficiencies, the model outperforms standard IRT-based methods on an online response prediction task when applied to real responses collected from students interacting with diverse pools of educational content. \end{abstract}

\section{Introduction}
\label{intro}
Accurately predicting student responses is an important task for adaptive learning systems. Better predictions enable more efficient diagnosis and remediation of student deficiencies. When reported as analytics, these predictions can also provide the student or teacher with actionable information to improve student outcomes. Traditional approaches mostly consider single-session assessments (e.g., a standardized examination), but to work in learning environments, adaptive systems must deal with significant challenges:

\begin{itemize}

\item
student knowledge states are constantly fluctuating, due to online or offline learning, forgetting, and varying levels of engagement with the content
\item
lack of control over the user experience: students may do different, unrelated, assignments on their own schedule, and at different paces
\item
student paths through content are often heavily correlated for pedagogical reasons (contrary to an ideal randomized controlled trial) and may include repeat attempts and/or be subject to selection bias
\item
up-to-date predictions must be available as soon as a new response is made (imagine a student aces a quiz and submits, then gets recommended similar material)
\item
accurate inferences and sensible recommendations must be made even at the beginning of a session, when little data about the student is available
\item
the pool of content is very diverse: fitting a model of student responses to one set of content may not generalize to another set.  Furthermore, relationships between content are difficult to establish and represent in the model, even given expert labels \cite{lindsey14}.
\end{itemize}

Bayesian Knowledge Tracing (BKT) is a framework that explicitly models student learning \cite{corbett95} but it accounts for diversity in neither students nor content. On the other hand, Item Response Theory (IRT) is a framework for modeling responses by estimating both student and content properties \cite{rasch60, lord80}, but it does not account for student learning. Several recent efforts have attempted to combine the complementary benefits of these two frameworks, with some success \cite{gonzalesbrenes13,khajah14,jascha13, lan14}. However, a few gaps remain: First, the way in which these models are evaluated is not consistent with the requirements of an adaptive learning system, which has access to all previous responses and is tasked with predicting the outcome of one or more prospective pairings of students and questions. Instead, model performance is gauged by predicting randomly held out chunks of responses (which is not a realistic task even if these responses are in the future and contiguous in time) for each student, or all responses for a randomly chosen subset of students (an approach that does not leverage currently available information). Second, the benefit of temporal methods over standard IRT methods has not been empirically observed. This could be due to high correlation in student paths through the content (observed in some of our data as well). Third, previous models either do not account for multiple dimensions of knowledge, or assume they can be modelled independently. Finally, data are typically collected via a highly instrumented and standardized system (intelligent tutoring system), which is unlike several other online learning environments.

We develop a unified IRT-based model that allows for student and item parameters, temporal fluctuation of student abilities, and content diversity. We compare this model with standard IRT models on an evaluation task identical to that performed in a production setting: predicting the next response given all previous responses to date. Our findings are twofold: First, incorporating temporality into our model yields a significant performance boost. Second, modeling multiple dimensions of knowledge along with the connections between them (based on labels from subject matter experts) provides  additional improvement in performance.

\section{Background}

\subsection{Knowledge tracing}

Knowledge tracing \cite{corbett95} (KT) is a framework for modeling the process of students mastering one or more skills while completing a sequence of assessments. At each time $t$, $Z_t$ is an unobserved binary variable indicating whether the student has mastered the skill or not, and $X_t$ is an observed binary variable indicating the correctness value of the completed assessment. A Hidden Markov Model is used to capture the statistical relationships between these. Emission probabilities govern how likely a correct response is given each hidden state. Transition probabilities govern how a student transitions between the hidden states. Several KT variants have been proposed to individually handle student abilities, multiple skills, assessment difficulty, and external aids (see \cite{khajah14} and the references therein). In \cite{gonzalesbrenes13,khajah14}, a unifying framework for these is proposed where the probabilities are parameterized in terms of these observed features.

\subsection{Item response theory}

Item response theory (IRT) \cite{lord80,rasch93} is a framework for modeling binary student responses on a set of assessments. In the two-parameter formulation the probability that a student $s$ answers an assessment $q$ correctly is given by:

\begin{equation}
p_{sq} = f(\alpha_q (\theta_s - \beta_q)) \label{eq:psq}
\end{equation}

where $\theta_s$ is the student proficiency, $\beta_q$ is the assessment difficulty, and $\alpha_q$ is the assessment discrimination which governs how sensitive the correctness probability is to the student’s proficiency. The function $f(.)$ is known as the \textit{item response function} and increases monotonically from 0 to 1. We choose $f(x) = \Phi(x)$, the cumulative distribution function of the normal distribution. This model is known as the two-parameter ogive, or 2PO model \cite{rasch93}.

An underlying assumption of this model is that student abilities ($\theta_s$) remain constant over time, making this more applicable to responses from an examination rather than a learning experience over an extended period of time. In \cite{gonzalesbrenes13} and \cite{jascha13}, the IRT framework is augmented to handle temporal fluctuations in student proficiencies. However, the benefits over standard IRT were not empirically realized in the former, while the latter required considerable amount of computing resources and time to fit.

Note also that none of the models in the KT or IRT families have been rigorously evaluated on the online prediction task.

\section{T-SKIRT}
%
%

\subsection{Temporality}

We extend the 2PO model by modeling the student proficiencies over time as a Wiener process. Under this model, the conditional relationships are given by

\begin{equation}
P(\theta_{t+\tau} | \theta_t ) = \phi_{\theta_t, \nu^2 \tau}(\theta_{t+\tau})
\end{equation}
where $\phi_{\mu, \sigma^2}(.)$ is the probability density function of the normal distribution with mean $\mu$ and variance $\sigma^2$. The proportionality constant $\nu^2$ on the conditional variance is a model parameter, tuned to maximize prediction accuracy on a separate dataset). Under this model, the joint probability over responses and proficiencies for a single student is given by:

\begin{eqnarray}
&&P(r_{1:t},\theta_{1:t}) = P(r_{1:t}|\theta_{1:t})P(\theta_{1:t}) \label{eq:joint} \\
&&= P(\theta_1)\prod_{i=2}^tP(\theta_i|\theta_{i-1})\prod_{i=1}^tP(r_i|\theta_i) \notag \\
&&= \phi_{\mu_0,\sigma_0^2}(\theta_1)\prod_{i=2}^t\phi_{\theta_{i-1},\nu^2}(\theta_i) \prod_{i=1}^tp_i^{r_i}(1-p_i)^{1-r_i} \notag
\end{eqnarray}

where $p_i = \Phi(\alpha_{q_i}(\theta_i-\beta_{q_i}))$ is the probability of correct on the $i'th$ response to question $q_i$, as per Eq.~\ref{eq:psq}. The primary task is to infer the posterior distribution over the current proficiencies $\theta_t$ given all past responses. We assume that item parameters have been learned previously and are fixed. Let $r_{1:n}$ and $\theta_{1:n}$ denote sequences of binary responses and real-valued proficiencies, respectively. The posterior is then given by:

\begin{equation}
P(\theta_t | r_{1:t}) \propto P(r_{1:t} | \theta_t) P(\theta_t) \label{eq:posterior}
\end{equation}

To evaluate the first term on the right-hand side, it is necessary to integrate out all possible paths $\theta_{1:t-1}$:

\begin{equation}
P(r_{1:t} | \theta_t) = \int P(r_{1:t}, \theta_{1:t-1} | \theta_{t})d\theta_{1:t-1} \notag
\end{equation}

However, computing this integral is expensive, especially in an online setting. For computational efficiency, we make the following approximation:

\begin{eqnarray}
&&P(r_{1:t} | \theta_t) \approx \prod_{i=1}^t P(r_i | \theta_t)  \notag \\
&&= \prod_{i=1}^t \int P(r_i, | \theta_i) P(\theta_i | \theta_t)d\theta_i \notag \\
&&=  \prod_{i=1}^t \int p_i^{r_i}(1-p_i)^{1-r_i} \phi_{\theta_t,\nu^2(t-i)}(\theta_i)d\theta_i \label{eq:int}
\end{eqnarray}

Using the definition of $p_i$ and the fact that $\int \Phi(\alpha(x-\beta))\phi_{\mu,\sigma^2}(x)dx = \Phi\left(\frac{\alpha(\beta - \mu)}{\sqrt{1+\alpha^2\sigma^2}}\right)$, we can simplify this to:
\begin{equation}
P(r_{1:t} | \theta_t) \approx \prod_{i=1}^t \tilde{p}_i^{r_i}(1-\tilde{p}_i)^{1-r_i} \label{eq:postapprox}
\end{equation}
where:
\begin{eqnarray}
\tilde{p}_i = \Phi(\tilde{\alpha}_i(\theta_t - \beta_{q_i})) \notag \\
\tilde{\alpha}_i = \frac{\alpha_{q_i}}{\sqrt{1 + \alpha_{q_i}^2\nu^2(t-i)}} \label{eq:effdisc}
\end{eqnarray}

Note that this formulation is exactly the same as the likelihood under a non-temporal IRT model, with $\alpha_{q_i}$ replaced by $\tilde{\alpha}_i$. Therefore, we refer to $\tilde{\alpha}_i$ as the ``effective discrimination'' of the $i$'th response on the student's current proficiency $\theta_t$. Note that the further back in time the response, the lower the effective discrimination and thus the smaller effect it has on the posterior over $\theta_t$ in Eq.~\ref{eq:posterior}. Finally, we can substitute Eq.~\ref{eq:postapprox} into Eq.~\ref{eq:posterior} and take the log to get the approximate log-posterior:

\begin{eqnarray}
\log P(\theta_t|r_{1:t}) &\approx& \log P(\theta_t) + \sum_{i=1}^t r_i\log(\tilde{p}_i)  \\
&+& (1-r_i)\log(\tilde{p}_i) \label{eq:logpost}
\end{eqnarray}

\subsection{Structured, multidimensional prior over student proficiencies}

When the content pool is very diverse, summarizing student proficiency with a single number (varying over time) may not be enough to capture the student response patterns. Content can be organized into groups against which student proficiency is measured. These groups have been referred to as knowledge components \cite{corbett97} or skills \cite{lindsey14,lan14}. The content used in our experiments has been grouped by subject matter experts into groups called “concepts.” Each student's proficiency at a single time is now represented by a vector of proficiencies in each concept, $\vec{\theta}_{t} \in \mathbb{R}^C$ where C is the total number of concepts. The IRT likelihood (Eq.~\ref{eq:psq}) can then be rewritten as:

\begin{equation}
p_{sq} = f(\alpha_q (\theta_{sc_q} - \beta_q)) \label{eq:psq_concepts}
\end{equation}

where $c_q$ is the concept assessed by question $q$.

Experts also identify prerequisite relationships between concepts indicating that content in one concept cannot be mastered without having mastered content in another concept. For more details see \cite{whitepaper}. These concepts and prerequisite relationships define a directed acyclic graph, and from this we can define a prior distribution over $\vec{\theta}_t$ that captures the conceptual relationships (similar to how expert labels were used in \cite{lindsey14}). We employ a specific multivariate Gaussian prior whose log-probability is given by:

\begin{equation}
\log P(\vec{\theta}_t) = -\lambda\sum_{n=1}^N\theta_{tn}^2 - \gamma \sum_{n,m:n\prec m}(\theta_n - \theta_m)^2
\label{eq:logpriormulti}
\end{equation}

where  $\theta_{tn}$ is the n'th element of the vector $\vec{\theta}_t$, $n \prec m$ indicates that concept $n$ is prerequisite to concept $m$, and the parameters $\lambda$ and $\gamma$ control the overall variance and relative correlations, respectively. Note that the precision  matrix has non-zero entries along the diagonal and only for pairs of concepts that have a prerequisite relationship. This defines a prior over proficiencies at any fixed point in time. In order to incorporate this into the temporal model, we make the following approximation for computational tractability:

\begin{equation}
P(\vec{\theta_i} | \vec{\theta_t}) \approx \prod_n P(\theta_{in} | \theta_{tn}) \label{eq:multiapprox}
\end{equation}

which allows us to have a multivariate analog of Eq.~\ref{eq:logpost}:

\begin{align}
\log P(\vec{\theta}_t|r_{1:t}) \approx& \log P(\vec{\theta}_t) + \sum_{i=1}^t r_i\log(\tilde{p}_i)  \\
&+ (1-r_i)\log(\tilde{p}_i)
\label{eq:logpostmulti}
\end{align}

Substituting Eq.~\ref{eq:logpriormulti} into Eq.~\ref{eq:logpostmulti} gives:
\begin{align}
\label{eq:logpostmultiexpanded}
\log P(\vec{\theta}_t|r_{1:t}) \approx&  -\lambda\sum_{n=1}^N\theta_{tn}^2 \\
&- \gamma \sum_{n,m:n\prec m}(\theta_n - \theta_m)^2\notag \\
&+ \sum_{i=1}^t r_i\log(\tilde{p}_i) + (1-r_i)\log(\tilde{p}_i) \notag
\end{align}

\section{Experimental setup}

Data was collected from a variety of educational products integrated with Knewton's adaptive learning platform and used in various classroom settings across the world. These products vary with respect to the educational content used (disciplines spanned math, science, and English language learning) as well as the way in which students are guided through the content. For example, students may take an initial assessment and then be remediated on areas needing improvement. In other products, student start from the beginning and work toward a predefined goal set by the teacher. In all of these settings, Knewton receives data about each interaction (anonymized student id, content module id, correctness, and timestamp).

\begin{figure}[ht]
\vskip 0.2in
\begin{center}
\centerline{\includegraphics[width=\columnwidth]{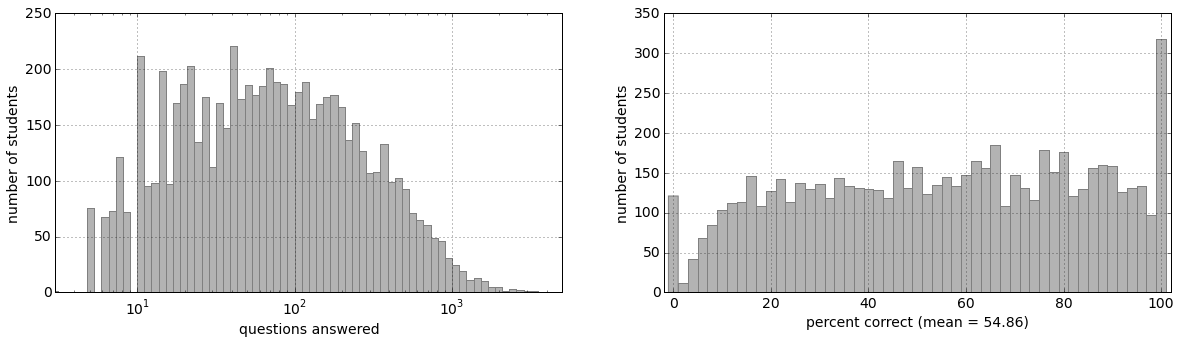}}
\caption{Data set student statistics.}
\label{fig:studentstats}
\end{center}
\vskip -0.2in
\end{figure}

We utilized approximately 1M responses of 6.3K randomly sampled students on 105.6K questions spanning roughly 4 months. Students who worked on fewer than 5 questions total were excluded. For each student/question pair, only the most recent 4 responses were used to avoid long strings of multiple attempts on questions. After pre-processing, student history lengths ranged from 5 to 3.2K responses, including repeat attempts (see Fig \ref{fig:studentstats}).  The overall percent correct of these responses is 54.6\%.

\begin{figure}[ht]
\vskip 0.2in
\begin{center}
\centerline{\includegraphics[width=\columnwidth]{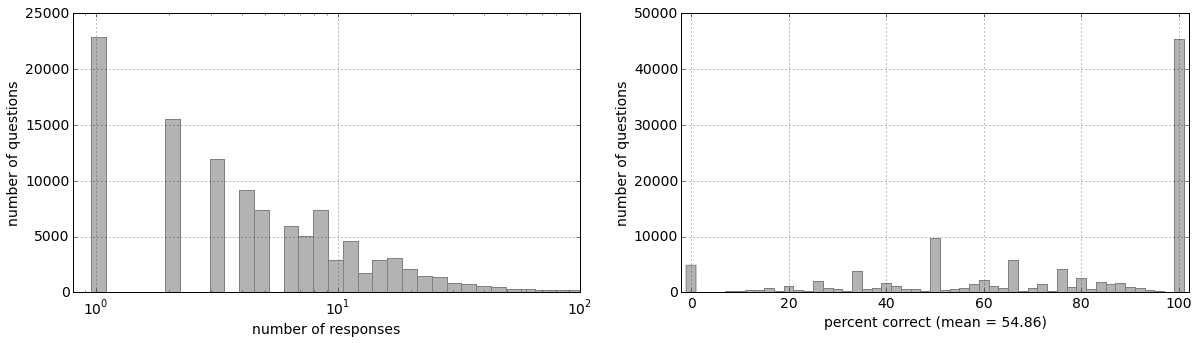}}
\caption{Data set item statistics.}
\label{fig:itemstats}
\end{center}
\vskip -0.2in
\end{figure}

\begin{figure*}[ht]
\vskip 0.2in
\begin{center}
\centerline{\includegraphics[width=\textwidth, trim=4cm 0cm 4cm 0cm]{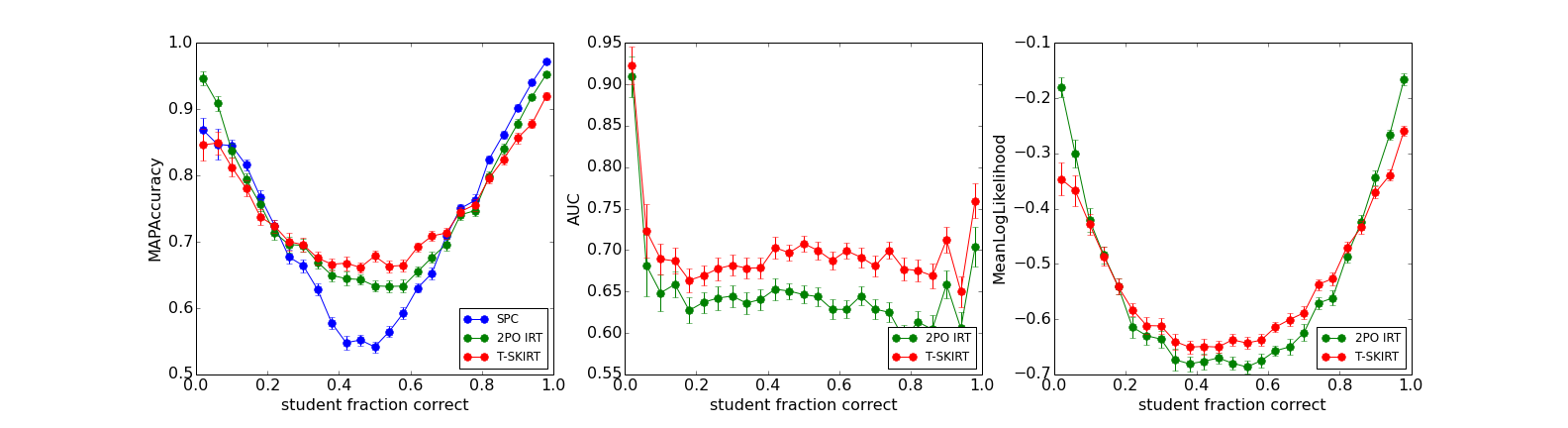}}
\caption{Performance of T-SKIRT (red) compared with running student percent-correct (blue) and 2PO IRT (green), as a function of the student's overall percent correct, with respect to prediction accuracy (left), AUC (middle), and average log-likelihood (right).}
\label{fig:acc_vs_pc}
\end{center}
\vskip -0.2in
\end{figure*}

\section{Results}

We compare the following student response models:

\begin{itemize}
\item
student percent-correct (SPC): predict correct if and only if the majority of previous responses for the student are correct.
\item
2PO IRT: standard ogive model, equivalent to Eq.~\ref{eq:logpost} with $\nu=0$ and $\lambda=1.0$.
\item
2PO temporal IRT: same as above but with $\nu=10.0$.
\item
Factorial MVN 2PO IRT: factorial multivariate Gaussian prior on proficiencies per concept (labelled by experts), using Eq.~\ref{eq:logpostmultiexpanded} with $\nu=0,\gamma=0$ and $\lambda=1.0$.
\item
Correlated MVN 2PO IRT: same as above but with $\gamma=0.5$.
\item
Correlated MVN temporal 2PO IRT (T-SKIRT): same as above but with $\gamma=0.5, \nu=0.1$.
\end{itemize}

To compare model performance in a task relevant for online adaptive learning systems, we measured the accuracy of predicting each student response given only the previous history of that student.  For the SPC model this amounts to predicting correct if the majority of previous interactions are correct. For models with latent student abilities, this entails estimating $\theta_t$ (or $\vec{\theta}_t$ for multi-dimensional proficiency models) by maximizing the logarithm of the (approximate) posterior probability (Eq.~\ref{eq:logpost} for single-dimensional, Eq.~\ref{eq:logpostmultiexpanded} for multi-dimensional). Note that both of these objective functions are convex w.r.t. the proficiencies and are easily optimized using first- or second-order gradient-based methods. We use this estimate to predict the correctness of the next interaction via Eq.~\ref{eq:psq_concepts}, and record the fraction of predictions that matched the observed responses.

The data was split into two parts. The first was used to estimate item difficulties ($\beta_q$'s) and discriminations ($\alpha_q$'s) using a standard 2PO IRT model with normal priors ($\beta_q \sim \mathcal{N}(0, 1), \alpha_q \sim \mathcal{N}(1,0.5)$). The parameters $\lambda,\beta,\nu^2$ were also tuned to optimize prediction accuracy on this data set. The second data set was used to evaluate and compare the models.

The results are summarized in Table.~\ref{tab:results}. Adding a temporal component to the standard 2PO IRT model (using the effective discriminations in Eq.~\ref{eq:effdisc}) yields a 2\% increase in accuracy. Using multidimensional proficiencies based on conceptual groupings identified by experts gives a 1.6\% increase. Adding information about conceptual relationships (using a correlated multivariate prior) brings this improvement to 1.9\%. Combining the multidimensional prior with the temporal component yields a total of 2.8\% improvement over standard 2PO IRT.

Figure~\ref{fig:acc_vs_pc} plots the improvement per student as a function of the student's overall fraction of correct responses, illustrating that the majority of the improvement comes from responses for students that are most difficult to predict. T-SKIRT does slightly worse than 2PO IRT for students who are doing very well or very poorly, and we are investigating why it underperforms in this regime.

\begin{tabular}{l c c}
\label{tab:results}
\textbf{Model} & \textbf{Accuracy $\pm$ 1 SEM} & AUC\\
SPC & 0.7085 $\pm$ 0.0021 & n/a \\
2PO IRT & 0.7201 $\pm$ 0.0018 & 0.7954 \\
2PO temporal IRT & 0.7420 $\pm$ 0.0018 & 0.8119 \\
Spherical MVN 2PO IRT & 0.7362 $\pm$ 0.0016 & 0.8056 \\
Correlated MVN 2PO IRT & 0.7390 $\pm$ 0.0016 & 0.8110 \\
T-SKIRT & 0.7478  $\pm$ 0.0016 & 0.8194
\end{tabular}

\section{Conclusions}

We developed a model, T-SKIRT, for predicting student responses that addresses two major challenges faced by an adaptive learning system: accounting for student learning and handling diversity in student ability and content properties. We evaluated this model on a task required in a production environment (predicting the next response of a student given all previous responses) and found that it gives superior predictions over standard IRT-based models when applied to real student data despite the large variability in content discipline, grade level, and learning environment. In contrast to \cite{gonzalesbrenes13}, incorporating temporality into the model yielded significant benefits over standard IRT. We also found that conceptual groupings and prerequisite relationships provided by experts also yielded significant improvement in contrast with \cite{lindsey14} where expert labels did not yield better predictions.

Our model can be improved or extended in several ways. For example, the temporal model can be augmented to account for observable events where learning is likely to occur (e.g., completing a lesson, using learning aids or hints). The model can also be extended to account for more gradual learning or forgetting of material by incorporating drift terms into the Wiener process prior. Another area of active research focuses on whether concept-module and inter-concept relationships can be automatically determined from data within this framework (e.g., \cite{lindsey14,lan14}) and encoded in a structured prior over student proficiencies. Finally, our inference method makes an approximation of conditional independence for the sake of tractability -- we find that this yields improved performance, but are also investigating the theoretical validity of this approximation.



\bibliography{icml_workshop_paper}
\bibliographystyle{icml2015}

\end{document}